\documentclass[journal]{IEEEtran}
\usepackage{graphicx}
\usepackage{subfigure}
\usepackage{multirow}
\usepackage{longtable}
\usepackage{algorithm}
\usepackage{algorithmic}
\usepackage{booktabs}
\usepackage{stfloats}
\usepackage{caption}
\usepackage{amssymb,amsmath}
\usepackage{color}
\usepackage{fancyheadings}
\usepackage{hyperref}
\usepackage{cite}

\begin{document}
\title{Fake face detection via adaptive manipulation traces extraction network}

\author{Zhiqing~Guo$^{1}$  \quad Gaobo~Yang$^{1}$ \quad Jiyou~Chen$^{1}$ \quad Xingming~Sun$^{2}$ \qquad \vspace{1pt}\\
$^{1}$Hunan University  \qquad $^{2}$Nanjing University of Information Science and Technology\qquad\qquad\\
\hspace{0.1in}{\tt\small \{guozhiqing, yanggaobo\}@hnu.edu.cn} \qquad {\tt\small cjyhn0302@gmail.com} \qquad  {\tt\small sunnudt@163.com} \\
}

\markboth{}%
{Shell \MakeLowercase{\textit{et al.}}: Fake face detection via adaptive manipulation traces extraction network}

\maketitle

\renewcommand{\headrulewidth}{0pt}

\begin{abstract}
With the proliferation of face image manipulation (FIM) techniques such as Face2Face and Deepfake, more fake face images are spreading over the internet, which brings serious challenges to public confidence. Face image forgery detection has made considerable progresses in exposing specific FIM, but it is still in scarcity of a robust fake face detector to expose face image forgeries under complex scenarios such as with further compression, blurring, scaling, etc. Due to the relatively fixed structure, convolutional neural network (CNN) tends to learn image content representations. However, CNN should learn subtle manipulation traces for image forensics tasks. Thus, we propose an adaptive manipulation traces extraction network (AMTEN), which serves as pre-processing to suppress image content and highlight manipulation traces. AMTEN exploits an adaptive convolution layer to predict manipulation traces in the image, which are reused in subsequent layers to maximize manipulation artifacts by updating weights during the back-propagation pass. A fake face detector, namely AMTENnet, is constructed by integrating AMTEN with CNN. Experimental results prove that the proposed AMTEN achieves desirable pre-processing. When detecting fake face images generated by various FIM techniques, AMTENnet achieves an average accuracy up to 98.52\%, which outperforms the state-of-the-art works. When detecting face images with unknown post-processing operations, the detector also achieves an average accuracy of 95.17\%.
\end{abstract}
\IEEEpeerreviewmaketitle

\section{INTRODUCTION}
\IEEEPARstart{f}{ace} images contain rich and intuitive personal identity information, which make them be commonly used for biometric authentication such as identifying individuals. However, face images also have vulnerability and weak privacy, which makes them easy to be forged. Especially over the last three years, tremendous progresses such as DeepFake, generative models \cite{GAN,glow,vae} and computer graphics (CG) based methods \cite{Face2Face} have made facial image manipulations (FIM) reach a photo-realistic level. This opens the door to a variety of face image applications such as interactive game, movie industry and photography. Nevertheless, FIM might be intentionally used for malicious purposes. In June 2019, the MIT Technology Review reported that the rapid spread of a doctored video, in which the White House speaker Nancy Pelosi was drunk, has frightened lawmakers in Washington\footnote{https://www.technologyreview.com/s/613676/deepfakes-ai-congress
-politics-election-facebook-social/.}. Similar AI-enhanced synthetic media are also likely to be used in serious scientific research. Apparently, these FIM techniques bring serious crisis to social security and public confidence.

\begin{figure}
  \centering
  \includegraphics[width=3.0in]{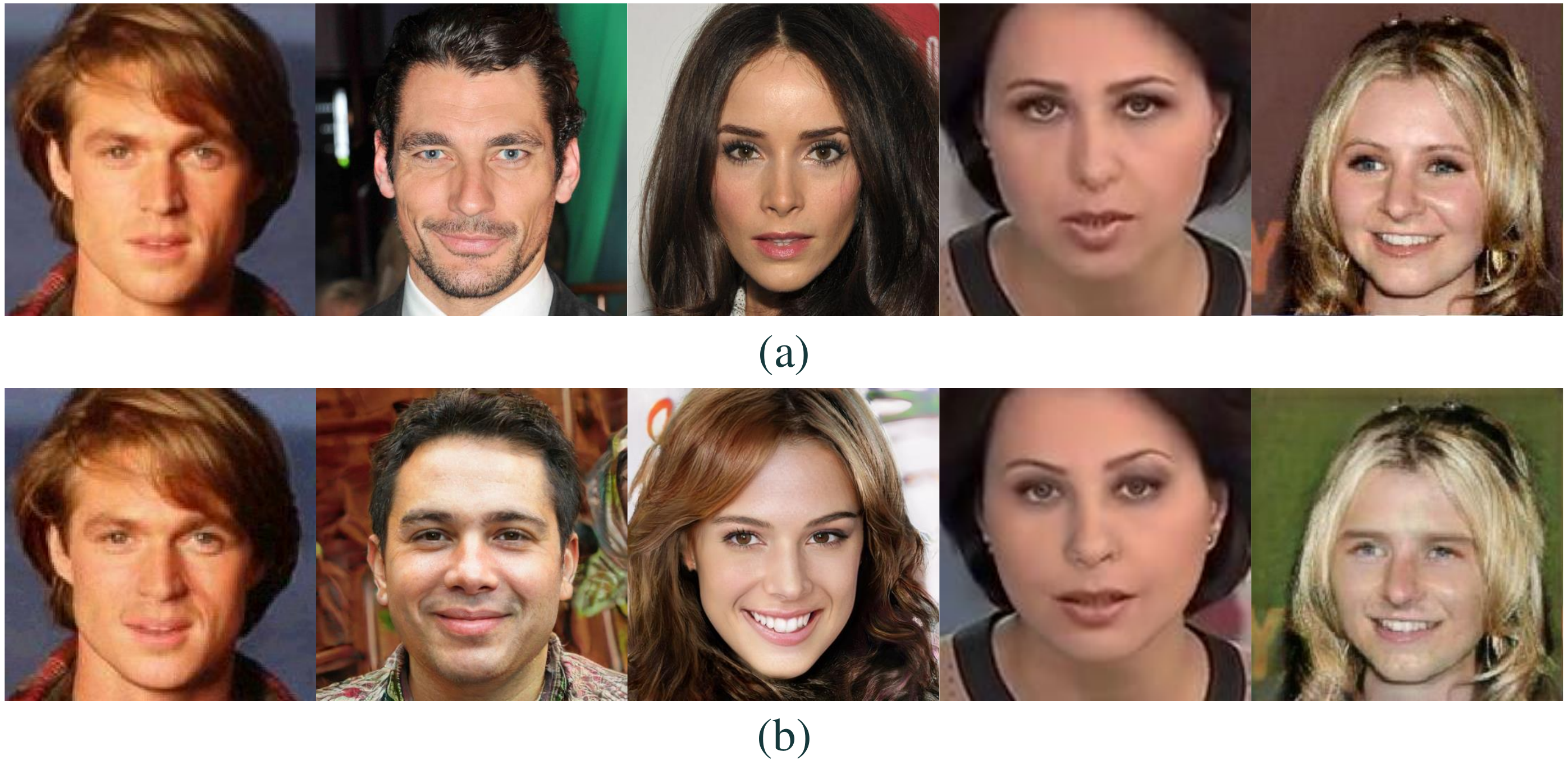}
  \caption{Can you identify which face image is fake? (a) Real face images with different resolutions. (b) From left to right, fake face images generated by Glow, StyleGAN, PGGAN, Face2Face, StarGAN, respectively.}\label{resultpic}
\end{figure}

Existing FIM techniques can be roughly divided into three categories: identity manipulation, expression manipulation and attribute transfer. Identity manipulation refers to generating fake face images of entirely imaginary people \cite{IntroVAE}, or replacing one face with the other one via FaceSwap \cite{FaceSwap} or DeepFake \cite{DeepFakes}.
Expression manipulation refers to generating face images with specific expressions \cite{glow}, or transferring facial expression from the source actor to the target face \cite{Face2Face}. For face attribute transfer, it refers to changing the styles of face images, such as age, gender, hair color, etc \cite{StarGAN}. In recent years, face identity manipulation has made great progress. The state-of-the-art works such as PGGAN \cite{pggan} and StyleGAN \cite{stylegan} can synthesize hyper-realistic fake face images with the resolution up to 1024$\times$1024.
Recent expression manipulation techniques also generate fake face images without leaving any perceptible artifacts. Several generative models including GANimation \cite{GANimation} and Glow \cite{glow} were proposed for expression manipulation with photo-realistic effects. Face2Face, which is a well-known Computer Graphics (CG)-based method, animates well the facial expression of the target video from a source actor \cite{Face2Face}. For face attribute transfer, there also exist some generative models such as StarGAN \cite{StarGAN}, and CycleGAN \cite{CycleGAN} which change facial attributes.
Fig. \ref{resultpic} shows some examples of face images. It is difficult for human eyes to differentiate those fake face images generated by various FIM techniques from real images. As we know, face image is an important form of non-verbal communication, from which we can perceive true information. When face images are maliciously manipulated, it might bring serious influence to people, especially politician and celebrities. Thus, face image forgery detection is becoming a key issue to be solved in the  community of image forensics.

Compared with the rapid progress of various FIM techniques, fake face image detection is lagging far behind. Most existing works were proposed to expose some specific FIM technique \cite{binary_guocc, binary_highpassfilter, binary_tdo, binary_lihaodong}, just providing binary classification about the trustworthiness of face images. Only a few works attempted to expose multiple FIM forgeries \cite{multi_generalization_wild}. Some works have studied the influence of post-processing \cite{MIPR,Capsule,Forensics2cozzolino2017}, yet they have not fully addressed fake face detection under complex scenarios.
Actually, face images are inevitably compressed or resized before spreading over social media, and possible post-processing operations might include JPEG compression (JP), Scaling (SC), Gaussian Blur (GB), Mean Filtering (ME) and Median Filtering (MED).
When the pre-trained detectors are detecting face images with unknown post-processing, there usually exist drastic performance degradations or they can be completely invalidated. Thus, the detection of multiple FIM forgeries under complex scenarios is becoming an urgent task to be solved. We need to develop a more general and robust fake face image detector.

The conventional image forensics framework is made up of feature extraction and classification \cite{forensics_framwork1, forensics_framwork2}. The extracted features are usually hand-crafted, which leads to poor generalization capability. In recent years, Convolutional Neural Network (CNN) has provided us an alternative for feature learning and classification automatically \cite{yu2018hierarchical}. Instead of learning content representation for image classification tasks, CNN should learn discriminative features from subtle manipulation traces for image forensics \cite{constrainedCNN}. Though CNN has achieved desirable accuracies when detecting image forgeries such as copy-move and JPEG recompression \cite{DL1Chen}, the existing CNN forms still have their own constraints. If we want to further improve detection accuracy, the convolution layer should be forced to learn features from tampering traces by improving its standard form. To the best of our knowledge, there is only one attempt, namely  MISLnet \cite{constrainedCNN}, to address this issue. Note that the first convolution layer, which is called the constrained convolution (Constrained-Conv) layer, extracts low-level manipulation traces for image forgery detection. In essence, the Constrained-Conv layer resets specific coefficients of the kernels after each iteration. Actually, since the extracted manipulation traces are fragile, they might be lost after passing through many layers. Though MISLnet provides some insights to improve the CNN model, there remain some open questions. First, is it the best way to reset some specific coefficients in the Constrained-Conv layer after each iteration? Second, can the low-level manipulation traces be reused to improve the performance of the model?

To address these questions, we are motivated to propose an Adaptive Manipulation Traces Extraction Network (AMTEN), which serves as pre-processing to suppress the side effects of image content. For existing image forensics works, the general pipeline is to predict manipulation traces and extract features from them for classification \cite{pip1popescu}. Instead, AMTEN outputs manipulation traces that can expose face image forgery, but the traces are obtained by subtracting the original image from the feature map. Note that AMTEN is different from ResNet \cite{ResNet}. ResNet generates an intermediate feature representation map by using the sum of the input feature map and residual block output, without extracting the manipulation traces. However, AMTEN extracts the manipulation traces by using the difference between the input image and the output feature map, which adapts well to capture the manipulation traces for tampering detection tasks. The extracted traces are reused in AMTEN to avoid information loss. A fake face detector, namely AMTENnet, is proposed to detect multiple FIM forgeries. The main works and contributions are summarized as follows.

\begin{itemize}
\item A pre-processing module, namely AMTEN, is specially designed for the CNN-based face image forensics. Quite different from the fixed predictors in existing works, AMTEN predicts manipulation traces adaptively during back-propagation. AMTEN provides more discriminative manipulation traces for face forensics tasks. Moreover, it might also serve as the basic manipulation traces predictor, which means that it can be transferred to the CNN-based models to detect other image forgeries.
\item By integrating AMTEN with CNN, a robust fake face detector, namely AMTENnet, is constructed to expose the state-of-the-art FIM forgeries under complex scenarios. To the best of our knowledge, this is the first attempt towards the detection of multiple FIM techniques.
\item By applying some post-processing operations including lossy compression, blurring and scaling to input images, we simulate practical face image forensics under complex scenarios as real as possible. To prove the effectiveness of the proposed AMTENnet, a series of experiments are conducted. It achieves higher detection accuracy than existing works. In addition, we explore the way to improve the generalization ability.
\end{itemize}

The remainder of this paper is organized as follows. Section 2 summarizes the related works on face image forensics. Section 3 presents the AMTENnet for fake face detection. Section 4 reports the experimental results and analysis, and we conclude in Section 5.

\section{Related Work}
Machine learning has been widely-used in fake face image detection \cite{LandmarkSVM}. To expose the face-swapping forgery, Zhang et al. \cite{BoW} constructed a feature set of bag of words, which provides distinguishable features into SVM for binary classification. To detect the Face2Face reenacted facial expression forgery in videos, Guo et al. \cite{binary_guocc} exploited both texture-based moment features and optical flow-based motion features. To expose the synthesized face images by GAN, Li et al. \cite{binary_lihaodong} defined a similarity index by Chi-square distance to model the disparities in color components. In addition, Agarwal et al. \cite{world_leader} proposed a deepfake face video detection approach by exploiting the correlation between facial expressions and movements. However, these machine learning based works usually have poor generalization capability.

Considering the particularity of face images, some methods exploited the biological inconsistency between real and fake faces. Li et al. \cite{eye_blinking} proposed to expose DeepFake videos by detecting the rate of eye blinking. Matern et al. \cite{visual_artifacts} exposed fake face images by exploiting some visual artifacts such as the defects of reflection details near eyes, and the imprecise geometry of both nose and teeth. Ciftci et al. \cite{FakeCatcher} presented a FakeCatcher to detect inauthentic portrait video by exploiting the biological signals of facial areas. Yang et al. \cite{head_poses} used 3D head poses to expose AI-generated fake face images. However, when there exist no obvious biological defects in fake face images, these methods might also be invalidated.

Some works have addressed face image forensics from new perspectives. Xuan et al. \cite{generalization_xuan} improved the detector's generalization ability by adding noises in the training stage. Cozzolino et al. \cite{ForensicTransfer} addressed the forensics transfer issue among different FIM techniques. Dang et al. \cite{ImbalanceData} considered the issue of imbalanced samples. Yu et al. \cite{Fingerprints} proposed to discriminate fake face images synthesized by different GANs by using their inherent fingerprints. Considering the time-consuming training of DeepFake, Li et al. \cite{SimulateArtifacts} simulated the DeepFake-generated negative samples via simple image post-processing such as GB.
Li et al. \cite{AdversarialPerturbations} also disrupted AI face synthesis with imperceptible adversarial perturbations. These efforts provide various insights to promote the development of face forensics towards universal forensics.

In recent years, CNNs have been widely used in image forensics tasks due to their superior performance \cite{ForensicTransfer}. These CNN models can be divided into three categories:
(1) Stacking standard CNN modules for a specific image forensics task \cite{binary_tdo,AppliedSciences};
(2) Using hand-crafted residual features extracted by either high pass filter or spatial rich model (SRM) for steganalysis, which are then input into CNN for image forensics \cite{binary_highpassfilter, SRMfilter};
(3) Improving the form of the convolution layer, such as the Constrained-Conv layer \cite{constrainedCNN}, to force CNN to directly learn features from tampering traces. However, the existing face image forensics works only
exploited either standard CNN or hand-crafted features for forensics. These works have no explicit restrictions on the convolution layer to learn features from tampering traces.
For example, Afchar et al. \cite{MesoNet} proposed a compact CNN model, namely MesoNet, for facial video forgery detection. It achieved an average detection accuracy up to 95\% on the FaceForensics dataset.
Dang et al. \cite{AppliedSciences} designed a customized CNN to detect face images generated by PGGAN \cite{pggan} and BEGAN \cite{BEGAN}. Li et al. \cite{FaceXray} proposed an image representation method, which can decompose the blending boundary of the face image to expose fake face images. Dang et al. \cite{Dang_attention} utilized the attention mechanism to highlight  manipulation region and improve feature representation for face forensics tasks.
Mo et al. \cite{binary_highpassfilter} introduced high pass filter into CNN to identify the PGGAN-generated faces. Gera et al. \cite{RNN} also proposed a temporal-aware pipeline to expose deepfake videos, which achieves an accuracy up to 97\%.
In general image forensics, MISLnet is the only work that learns features from tampering traces by limiting convolution layer. Inspired by MISLnet, we design a specifical pre-processing module based on the convolution layers, namely AMTEN, to predict manipulation traces. We also propose a robust fake face detector, namely AMTENnet, to learn discriminative features from manipulation traces.

\begin{figure*}
  \centering
  \includegraphics[width=5.5in]{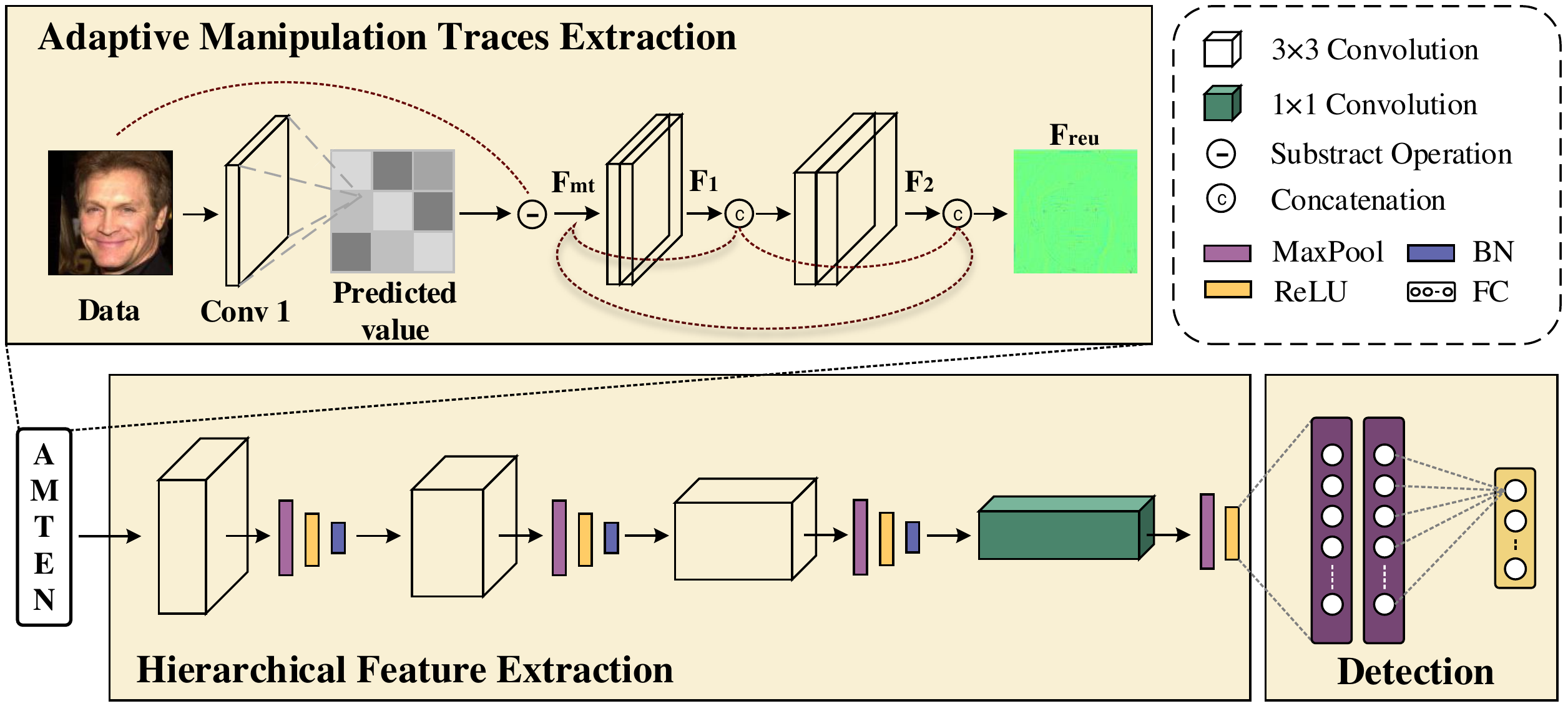}
  \caption{The proposed AMTENnet architecture. Given an input RGB image, we use the Conv 1 in AMTEN to obtain the feature map of image. Then, the original image is subtracted from the feature map in Conv 1 to extract the low-level manipulation traces $F_{mt}$. Furthermore, the stable higher-level manipulation traces, namely $F_{reu}$, are obtained by reusing the $F_{mt}$. Next, the $F_{reu}$ are passed to the subsequent convolution layers for hierarchical feature extraction to obtain high-level forensics features. Finally, we use fully connected layers and softmax function to classify the images. MaxPool: Max Pooling Layer; ReLU: Rectified Linear Unit; BN: Batch Normalization; FC: Fully Connected Layer; $F_1$ and $F_2$ represent the feature maps of the previous layer, respectively.}\label{architecture}
\end{figure*}

\section{Proposed AMTENnet Model}
\subsection{AMTEN}
Some existing works such as SRM \cite{Forensics6fridrich2012} and SPAM \cite{SPAM} learn features from predicted manipulation traces. They firstly generate a set of predicted pixel value via a fixed predictor $f(\cdot)$. Then, the manipulation traces $T$ is obtained by subtracting the original pixel value from the predicted pixel value. That is,
\begin{equation}\label{eq_1}
     T = f(I) - I
\end{equation}
where $I$ is the input image. Then, the manipulation traces are used as low-level features to construct high-level features for image forensics.

\begin{figure*}
  \centering
  \includegraphics[width=6.0in]{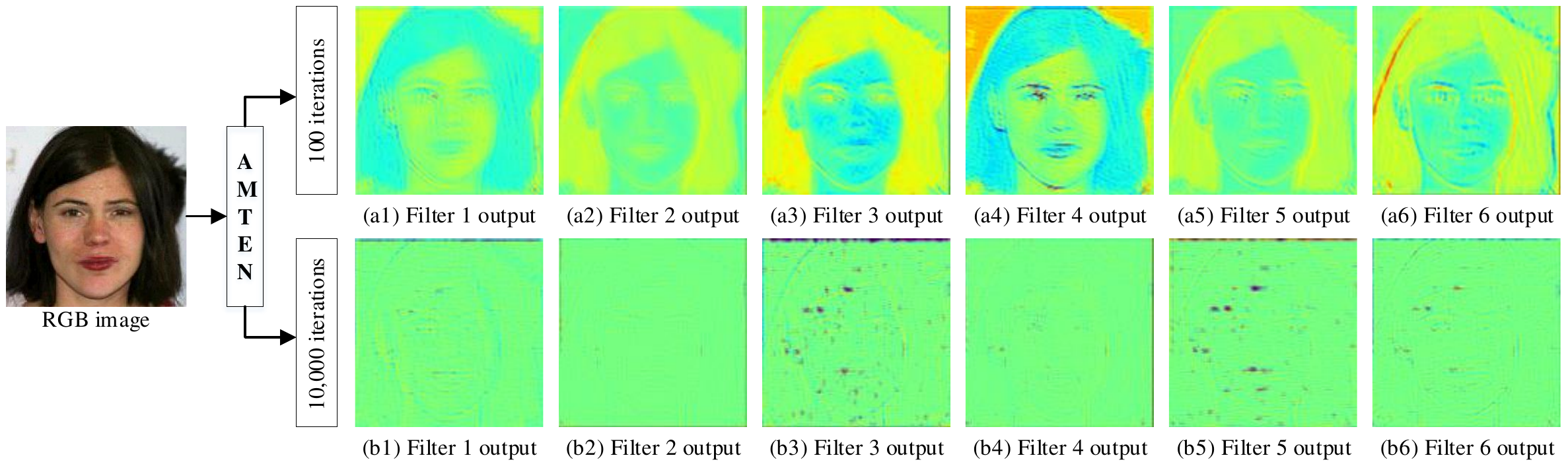}
  \caption{The output of the six filters in AMTEN at different iterations. As the number of iterations increases, the feature map obtained by the AMTEN gradually suppresses the image content and retains the trace features.}\label{AMTENresult}
\end{figure*}

To mimic this pipeline, AMTEN is specifically designed to automatically learn manipulation traces. Then, CNN is used to learn high-level features from manipulation traces due to its strong feature representation capability. As shown in Fig. \ref{architecture}, the first convolution layer (Conv 1) is used to predict the pixel value as follows.
\begin{equation}\label{eq_2}
     F_{j}=\sum_{i=1}^{m}I_{i}\ast\omega_{ij}+b_{j}
\end{equation}
where $F_{j}$ is the $j^{th}$ feature map which is output by the Conv 1 layer, $I_{i}\ast\omega_{ij}$ represents the convolution between the $i^{th}$ channel of the input image $I$ and the $i^{th}$ channel of the $j^{th}$ convolutional kernel in Conv 1, and $b_{j}$ is the bias term of the $j^{th}$ convolutional kernel. Then, the manipulation traces $F_{mt}$ are obtained by
\begin{equation}\label{eq_3}
  F_{mt}=F_{j}-I
\end{equation}
Apparently, the way to obtain manipulation traces in Equation (3) is almost the same as Equation (1). We randomly initialize the coefficients of the Conv 1 layer. Then, the weights are updated by an iterative algorithm during the back-propagation pass. In this paper, we adopt stochastic gradient descent (SGD) to train the model. The rules for iterative updates are defined as follows.
\begin{equation}\label{eq_4}
     \bigtriangledown \omega_{ij}^{(n)}=\varepsilon \frac{\partial E}{\partial \omega_{ij}^{(n-1)}}-\theta_1\cdot\bigtriangledown\omega_{ij}^{(n-1)}+\theta_2\cdot\varepsilon\cdot\omega_{ij}^{(n-1)}
\end{equation}
\begin{equation}\label{eq_5}
     \omega_{ij}^{(n)}=\omega_{ij}^{(n-1)}-\bigtriangledown \omega_{ij}^{(n)}
\end{equation}
\noindent where $\bigtriangledown$ is the gradient, $\omega_{ij}^{(n)}$ is the weight of the $i^{th}$ channel of the $j^{th}$ convolutional kernel in the $n^{th}$ layer, $\varepsilon$ is the learning rate and $E$ is the loss function. We use the momentum $\theta_1$ and the decay $\theta_2$ to accelerate model training. In the iterative training process, its goal is to minimize the average loss $E$ between true label and network output to make it converge. AMTEN iteratively adjusts the weights to obtain better manipulation traces. The average loss $E$ is defined as
\begin{equation}\label{eq_6}
     E=-\frac{1}{x} \sum_{i=1}^{x}\sum_{k=1}^{p} L_{i}^{(k)}log(y_{i}^{(k)})
\end{equation}
where $L_{i}^{(k)}$ is the true label of the $i^{th}$ image in the $k^{th}$ class, $y_{i}^{(k)}$ is the network output, $x$ is the number of training sample, and $p$ is the number of neurons in the output layer. In binary classification, $p$ = 2, which corresponds to real face image and fake face image respectively. In the multiple classification, $p$ represents the number of categories, which more accurately corresponds to the category to which each type of image belongs.

However, the manipulation traces $F_{mt}$ extracted by the Conv 1 layer are fragile. If they are used directly, it might still lead to unstable training. To address this issue, we borrow the idea of feature reusing from DenseNet \cite{DenseNet}. Let $c_1$ and $c_2$ be two convolution layers, and $H_{c_1,c_2}(\cdot)$ denote the composite function of $c_1$ and $c_2$. Let $[\beta_1,\beta_2,...,\beta_n]$ be the concatenation of the $n$ feature maps. Thus, the obtained traces $F_{mt}$ is passed into Conv 2 and Conv 3 to obtain the intermediate feature map as follows.
\begin{equation}\label{eq_7}
  F_1=H_{c_2,c_3}(F_{mt})
\end{equation}
Then, the feature map obtained by concatenating $F_1$ and $F_{mt}$ is passed to Conv 4 and Conv 5, which can be expressed as
\begin{equation}\label{eq_8}
  F_2=H_{c_4,c_5}([F_1,F_{mt}])=H_{c_4,c_5}([H_{c_2,c_3}(F_{mt}),F_{mt}])
\end{equation}
Finally, stable manipulation traces are obtained as follows.
\begin{equation}\label{eq_9}
  F_{reu}=[F_2,F_{mt},[H_{c_2,c_3}(F_{mt}),F_{mt}]]
\end{equation}

From AMTEN, stable manipulation traces are obtained by suppressing image content. Fig. \ref{AMTENresult} compares the manipulation traces obtained by different filters after 100 and 10,000 times of iterations. From it, we can observe that when AMTEN iterates 100 times, most image contents are not suppressed. However, when the iteration times reach 10,000, most image contents are suppressed whereas keeping well manipulation traces. Note that different from the fixed predictor in existing works, AMTEN adaptively learns manipulation traces, which are more suitable for face image forensics.

\subsection{Network Architecture}
For the proposed AMTENnet, most of the convolution layers adopt 3$\times$3 kernels, since it has been claimed that the 3$\times$3 convolutional kernel outperforms larger kernels \cite{vggnet}. Because there are three color channels in the input images, the Conv 1 layer uses three convolutional kernels to obtain the feature maps, respectively. Then, the manipulation traces $F_{mt}$ are obtained by subtracting the above feature maps from the original image. Let the input image be of size 128$\times$128. As shown in Fig. \ref{architecture}, $F_1$ and $F_{mt}$ are firstly concatenated to obtain feature maps, whose dimension is 128$\times$128$\times$6. To fully exploit the features of the previous layer, the number of the convolutional kernels in the successive layer should not be less than the number of channels of the input feature map. Thus, six convolutional kernels are used for Conv 4 and Conv 5. Table \ref{tab1} summarizes the parameters of AMTENnet. In Section 4.3, we will also analyze the influence of the number of convolutional kernels.

AMTEN obtains desirable manipulation traces $F_{reu}$. Instead of directly using them as the features for face image forensics, we design a hierarchical feature extraction (HFE) module. $F_{reu}$ is fed into the HFE module to learn high-level features. Specifically, the HFE module is made up of four convolution layers, four max-pooling layers (MaxPool), four ReLU activation functions (ReLU), and three batch normalization (BN) layers.

For the four convolution layers, we gradually increase the number of the convolutional kernels. That is, Conv 6=24, Conv 7=48, Conv 8=64, and Conv 9=128. For the convolutional kernels, small stride can extracts more abundant features than large stride. Thus, the stride of each convolution layer is set to 1. Before feeding the feature maps into the classification module, another convolution layer, namely Conv 9, is introduced to achieve cross-channel interaction. Different from the previous convolution layers, the Conv 9 layer adopts a 1$\times$1 kernel. It learns the linear combination of those features located in the same location but different channels.

Each convolution layer is followed with other types of layers, which include MaxPool, ReLU and BN. The MaxPool layer retains the most representative information (i.e., the maximum value) within the sliding window. It also reduces the dimension of feature maps, and introduces network nonlinearity to prevent over-fitting. For the four MaxPool layers, they use the same kernel size of 3$\times$3. To reduce the dimension of feature maps, the stride of each MaxPool layer is set to 2. The ReLU layer is introduced to increase network nonlinearity and overcome gradient vanishing \cite{ReLU}. Thus, the AMTENnet model can approximate any nonlinear function. Note that these nonlinear operations including MaxPool and ReLU are not introduced into AMTEN, which prevents the learned manipulation traces from being destroyed by them. To accelerate training, the BN layer is also used  in the AMTENnet model to regularize the output of the convolution layers.

Finally, the learned features are passed into the classification module, which is made up of three fully connected layers. The first two fully connected layers, which learn the associations among the deep features, have 300 neurons, respectively. The neurons in the last fully connected layer, whose outputs correspond to the real face image and possible face image manipulations.

\begin{table}[]
\caption{\label{tab1}Specification of the AMTENnet. ``conv", ``maxpool" and ``fc" correspond to the variables in Fig. 2}
\centering
\resizebox{55mm}{!}{
\begin{tabular}{|c|c|c|c|c|}
\hline
\multicolumn{5}{|c|}{Configuration} \\ \hline
Layers & Kernel sizes & Kernel quantities & Strides & Output sizes \\ \hline
Conv 1 & 3$\times$3 & 3 & 1 & 128$\times$128 \\ \hline
Conv 2 & 3$\times$3 & 3 & 1 & 128$\times$128 \\ \hline
Conv 3 & 3$\times$3 & 3 & 1 & 128$\times$128 \\ \hline
Conv 4 & 3$\times$3 & 6 & 1 & 128$\times$128 \\ \hline
Conv 5 & 3$\times$3 & 6 & 1 & 128$\times$128 \\ \hline
Conv 6 & 3$\times$3 & 24 & 1 & 128$\times$128 \\ \hline
MaxPool & 3$\times$3 & \textbackslash & 2 & 64$\times$64 \\ \hline
Conv 7 & 3$\times$3 & 48 & 1 & 62$\times$62 \\ \hline
MaxPool & 3$\times$3 & \textbackslash & 2 & 31$\times$31 \\ \hline
Conv 8 & 3$\times$3 & 64 & 1 & 29$\times$29 \\ \hline
MaxPool & 3$\times$3 & \textbackslash & 2 & 14$\times$14 \\ \hline
Conv 9 & 1$\times$1 & 128 & 1 & 14$\times$14 \\ \hline
MaxPool & 3$\times$3 & \textbackslash & 2 & 7$\times$7 \\ \hline
\multicolumn{5}{|c|}{FC 1: 300-dimension} \\ \hline
\multicolumn{5}{|c|}{FC 2: 300-dimension} \\ \hline
\multicolumn{5}{|c|}{Softmax Function} \\ \hline
\end{tabular}
}
\end{table}

\section{Experimental results and analysis}
\subsection{Experimental Settings}
Most existing works provide only binary classification about the trustworthiness of face images, without considering more complex scenarios in practical forensics. In the experiments, we simulate some complex scenarios as real as possible to verify the effectiveness of the proposed AMTENnet model. We conduct four groups of experiments.
First, AMTENnet is used to detect multiple FIMs. Second, we discuss the design of the AMTENnet model. Third, some comparisons are made among AMTENnet and the state-of-the-art works. Fourth, we explore the way to improve the robustness of AMTENnet in complex scenarios.

\textbf{Datasets.} To conduct the above experiments, we firstly build a hybrid fake face (HFF) dataset \footnote{https://github.com/EricGzq/Hybrid-Fake-Face-Dataset}, which contains eight types of face images. For real face images, three types of face images are randomly selected from three open datasets. They are low-resolution face images from CelebA \cite{celeba}, high-resolution face images from CelebA-HQ \cite{pggan}, and face video frames from FaceForensics \cite{Faceforensics}, respectively. Thus, real face images under internet scenarios are simulated as real as possible. Then,
some most representative FIM techniques, which include PGGAN and StyleGAN for identity manipulation, Face2Face and Glow for face expression manipulation, and StarGAN for face attribute transfer, are selected to produce fake face images. Note that since StarGAN can transfer facial attributes such as hair color (black, blond, brown) gender (male or female) and age (young or old) to other domains, five types of face attributes are manipulated via StarGAN. It has been claimed that face images with different attributes share the same artifacts or fingerprints when they are generated by the same GAN \cite{Fingerprints}. We mark these images with different attributes as StarGAN-generated. Table \ref{tab2} summarizes the details of the HFF dataset.
In addition, the open FaceForensics++ (FF++) dataset is used for experiments. There are 1k original video sequences, which are manipulated by four FIM techniques including Deepfakes \footnote{https://github.com/deepfakes/faceswap.}, Face2Face \cite{Face2Face}, FaceSwap \cite{FaceSwap} and NeuralTextures \cite{NeuralTextures}. The tampered videos are further compressed with two quality levels, namely high quality videos (HQ) and low quality videos (LQ).
In our experiment, we extract 50k and 10k face images from real video sequences as training sets and testing sets. For four kinds of tampered videos, we extract 12.5k and 2.5k face images from each category as training sets and testing sets. Note that to facilitate the experiment, we only exploit two compressed versions of the FF++ dataset in the experiment. That is, the HQ and LQ datasets contain 100k face images for training and 20k face images for testing, respectively.

\begin{table}[]
\caption{\label{tab2}The details of HFF datasets}
\resizebox{85mm}{!}{
\begin{tabular}{|c|c|l|c|c|}
\hline
\multicolumn{1}{|l|}{}            & Data Type            & \multicolumn{1}{c|}{Description}                                                                                      & Image Size       & Corpus Size \\ \hline
\multirow{3}{*}{Real Face Images} & CelebA               & Low-resolution face images                                                                                            & 178$\times$218   & 25k         \\ \cline{2-5}
                                  & CelebA-HQ            & High-resolution face images                                                                                           & 1024$\times$1024 & 10k         \\ \cline{2-5}
                                  & YouTube-Frame & Face video frames                                                                                                     & Random size      & 25k         \\ \hline
\multirow{5}{*}{Fake Face Images} & PGGAN                & \multirow{2}{*}{\begin{tabular}[c]{@{}l@{}}A generative model based \\ identity manipulation technique.\end{tabular}} & 1024$\times$1024 & 10k         \\ \cline{2-2} \cline{4-5}
                                  & StyleGAN             &                                                                                                                       & 1024$\times$1024 & 10k         \\ \cline{2-5}
                                  & Glow                 & \begin{tabular}[c]{@{}l@{}}A generative model based \\ expression manipulation technique.\end{tabular}                & 256$\times$256   & 25k         \\ \cline{2-5}
                                  & Face2Face            & \begin{tabular}[c]{@{}l@{}}A CG-based expression \\ manipulation technique.\end{tabular}                              & Random size      & 25k         \\ \cline{2-5}
                                  & StarGAN              & \begin{tabular}[c]{@{}l@{}}A generative model based \\ attribute transfer technique.\end{tabular}                     & 256$\times$256   & 25k         \\ \hline
\end{tabular}
}
\end{table}

\textbf{Evaluation Criterion.}
For image forgery detectors, they are usually evaluated by classification accuracy. In our tasks, since the distribution of data is roughly balanced, we also use classification accuracy for performance evaluation. To further evaluate the performance gains brought by different AMTEN design selections, the relative error reduction (RER) is also used as performance evaluation metrics. Let $E_1$ and $E_2$ be the numbers of errors for two detectors ($E_1>E_2$). $\mathrm{RER}$ is defined as $\mathrm{RER}=(E_1-E_2)/E_1$.

\textbf{Baseline Models.}
We choose some state-of-the-art works as the baselines for making experimental comparisons. They are summarized below.
\begin{itemize}
  \item Meso-4 \cite{MesoNet}: It exploits the mesoscopic properties of face images for facial forgery detection.
  \item MesoInception-4 \cite{MesoNet}: It is an improved version of Meso-4 and has better image forgery detection performance.
  \item Hand-Crafted-Res \cite{binary_highpassfilter}: Three high pass filters are used as pre-processing to extract hand-crafted features. The parameters with the best performance are used for comparisons.
  \item MISLnet \cite{constrainedCNN}: It exploits the Constrained-Conv layer, which suppresses image content and adaptively learns low-level residual features for universal forensics.
  \item XceptionNet \cite{FaceForensics++}: For the FF++ dataset, XceptionNet achieved the best performance.
  \item Model-base: To prove the gains brought by AMTEN, it is removed from the AMTENnet model. The remaining network is called the Model-base here.
  \item Hand-Crafted-Res-Model-base: We use the hand-crafted feature extractor in \cite{binary_highpassfilter} to replace the proposed AMTEN.
  \item Constrained-Conv-Model-base: AMTEN is replaced by the Constrained-Conv in \cite{constrainedCNN}.
  \item SRM-Model-base: AMTEN is replaced by the SRM filter kernels in \cite{SRMfilter}.
\end{itemize}

\textbf{Implementation Details.} The proposed AMTENnet model is implemented under the Caffe framework. We convert all face images in the datasets into the LMDB format and then resize them into 128$\times$128 for use in Caffe.
Each forensics model has 10 training epochs. We record the detection accuracies on the testing set after every 1000 iterations. Two Nvidia GeForce GTX 1080 Ti GPUs are used to train the model.

\begin{table}[]
\caption{\label{Confusion model-base}Confusion matrix for identifying various types of manipulations using model-base. The asterisks ``*" represents the value are below 1\%.}
\centering
\resizebox{90mm}{!}{
\begin{tabular}{|c|c|c|c|c|c|c|c|c|c|}
\hline
 & \multicolumn{9}{c|}{\textbf{Predicted class}} \\ \hline
\multirow{9}{*}{\textbf{The class}} &  & \textbf{CelebA} & \textbf{CelebA-HQ} & \textbf{YouTube-Frame} & \textbf{Glow} & \textbf{StarGAN} & \textbf{PGGAN} & \textbf{StyleGAN} & \textbf{Face2Face} \\ \cline{2-10}
 & \textbf{CelebA} & \textbf{99.48\%} & * & * & * & * & * & * & * \\ \cline{2-10}
 & \textbf{CelebA-HQ} & * & \textbf{87.05\%} & * & * & * & 11.90\% & * & * \\ \cline{2-10}
 & \textbf{YouTube-Frame} & * & * & \textbf{92.28\%} & * & * & * & * & 7.62\% \\ \cline{2-10}
 & \textbf{Glow} & * & * & * & \textbf{99.90\%} & * & * & * & * \\ \cline{2-10}
 & \textbf{StarGAN} & * & * & * & * & \textbf{99.74\%} & * & * & * \\ \cline{2-10}
 & \textbf{PGGAN} & * & 19.35\% & * & * & * & \textbf{80.25\%} & * & * \\ \cline{2-10}
 & \textbf{StyleGAN} & * & * & * & * & * & * & \textbf{99.80\%} & * \\ \cline{2-10}
 & \textbf{Face2Face} & * & * & 1.98\% & * & * & * & * & \textbf{97.98\%} \\ \hline
\end{tabular}
}
\end{table}

\begin{table}[]
\centering
\caption{\label{Confusion AMTENnet}Confusion matrix for identifying various types of manipulations using AMTENnet. The asterisks ``*" represents the value are below 1\%.}
\resizebox{90mm}{!}{
\begin{tabular}{|c|c|c|c|c|c|c|c|c|c|}
\hline
 & \multicolumn{9}{c|}{\textbf{Predicted class}} \\ \hline
\multirow{9}{*}{\textbf{The class}} &  & \textbf{CelebA} & \textbf{CelebA-HQ} & \textbf{YouTube-Frame} & \textbf{Glow} & \textbf{StarGAN} & \textbf{PGGAN} & \textbf{StyleGAN} & \textbf{Face2Face} \\ \cline{2-10}
 & \textbf{CelebA} & \textbf{99.56\%} & * & * & * & * & * & * & * \\ \cline{2-10}
 & \textbf{CelebA-HQ} & * & \textbf{95.30\%} & * & * & * & 3.45\% & * & * \\ \cline{2-10}
 & \textbf{YouTube-Frame} & * & * & \textbf{97.68\%} & * & * & * & * & 2.10\% \\ \cline{2-10}
 & \textbf{Glow} & * & * & * & \textbf{99.92\%} & * & * & * & * \\ \cline{2-10}
 & \textbf{StarGAN} & * & * & * & * & \textbf{99.66\%} & * & * & * \\ \cline{2-10}
 & \textbf{PGGAN} & * & 8.50\% & * & * & * & \textbf{91.45\%} & * & * \\ \cline{2-10}
 & \textbf{StyleGAN} & * & * & * & * & * & * & \textbf{99.85\%} & * \\ \cline{2-10}
 & \textbf{Face2Face} & * & * & * & * & * & * & * & \textbf{99.38\%} \\ \hline
\end{tabular}
}
\end{table}

\subsection{Detection of multiple FIM forgeries}
Firstly, the proposed detector is used to expose multiple FIM techniques simultaneously. That is, each type of face images are randomly selected from the HFF dataset and divided into three sub-datasets for training (75\%), validation (5\%), and testing (20\%), respectively. Note that test images have never appeared in the training set and the validation set. In the experiments, there are about 116k face images for training, which include real images with different resolutions and five types of fake images. When training the AMTENnet model, SGD is used for iterative optimization, and we set the momentum $\theta_1 = 0.95$ and the decay $\theta_2 = 0.005$. The learning rate is defined as
\begin{equation}\label{eq_10}
     \varepsilon = \varepsilon_{b} \times \gamma^{\lfloor(\frac{\alpha}{N})\rfloor}
\end{equation}
where $\varepsilon_{b}$ is the basic learning rate, $N$ is the fixed step size, $\alpha$ denotes current iteration, and $\lfloor\rfloor$ denotes rounding down. Their initial values are as follows: $\varepsilon_{b} = 0.001$, $\gamma=0.5$, $N = 1000$. With the increment of the iteration times, $\varepsilon$ decreases periodically. The batch size is set to 64. Each training epoch requires 1,817 iterations.

The confusion matrixes of Model-base and AMTENnet are reported in Table \ref{Confusion model-base} and Table \ref{Confusion AMTENnet}, respectively. Their average detection accuracies are 96.16\% and 98.52\%, respectively. We can observe from Table \ref{Confusion model-base} that the false detection rate between PGGAN and CelebA-HQ is high. Actually, two types of face images share similar textures. They are difficult to be detected, especially when they are resized into 128$\times$128. For Face2Face and YouTube-Frame, there also exist the same phenomenon. Note that when AMTEN is added into Model-base, which turns into AMTENnet, the false detection rate is greatly reduced.

Some comparisons are also made among different residual extraction methods, which include Hand-Crafted-Res, SRM, Constrained-Conv and AMTEN. In the experiments, they are followed with the same basic CNN model (Model-base). Table \ref{residual comparison} compares the experimental results achieved by them. From it, AMTENnet achieves the highest accuracy of 98.52\%, which proves that AMTEN achieves the best residual extraction. The reasons behind this are summarized as follows. Both SRM and Hand-Crafted-Res use the fixed filter to extract manipulation traces, Constrained-Conv resets specific coefficients after each iteration, whereas AMTEN adaptively updates the coefficients during the back-propagation pass to predict manipulation traces. Furthermore, AMTEN introduces feature reusing to improve detection accuracy.

\begin{table}[]
\centering
\caption{\label{residual comparison}The comparison of different residual extraction methods.}
\resizebox{45mm}{!}{
\begin{tabular}{|l|c|}
\hline
\multicolumn{1}{|c|}{Methods} & Accuracy         \\ \hline
Hand-Crafted-Res-Model-base   & 97.50\%          \\ \hline
Constrained-Conv-Model-base   & 95.24\%          \\ \hline
SRM-Model-base                & 97.49\%          \\ \hline
AMTENnet                        & \textbf{98.52\%} \\ \hline
\end{tabular}
}
\end{table}

\begin{table}[]
\centering
\caption{\label{different models}Identification rate for different CNN models.}
\resizebox{80mm}{!}{
\begin{tabular}{|c|c|c|c|c|}
\hline
\multicolumn{2}{|c|}{Models} & Description & \begin{tabular}[c]{@{}c@{}}Average\\ Accuracy\end{tabular} & RER \\ \hline
\multicolumn{2}{|c|}{AMTENnet} & \textbackslash{} & \textbf{98.52\%} & - \\ \hline
\multicolumn{2}{|c|}{Model-base} & Remove the AMTEN & 96.16\% & 61.46\% \\ \hline
\multirow{6}{*}{Modified} & AMTEN\_1 & \begin{tabular}[c]{@{}c@{}}Image data is used instead of\\ residual features $F_{mt}$\end{tabular} & 97.14\% & 48.25\% \\ \cline{2-5}
 & AMTEN\_2 & $F_{mt}$ is not reused & 97.46\% & 41.73\% \\ \cline{2-5}
 & AMTEN\_3 & Conv 4 = 3 and Conv 5 = 3 & 97.42\% & 42.64\% \\ \cline{2-5}
 & AMTEN\_4 & Conv 4 = 12 and Conv 5 = 12 & 97.88\% & 30.19\% \\ \cline{2-5}
 & AMTEN\_5 & \begin{tabular}[c]{@{}c@{}}5$\times$5 convolutional\\ kernel as the predictor\end{tabular} & 97.81\% & 32.42\% \\ \cline{2-5}
 & AMTEN\_6 & Remove Conv 3 and Conv 5 & 98.29\% & 13.45\% \\ \hline
\multirow{2}{*}{Other} & AMTENnet\_7 & \begin{tabular}[c]{@{}c@{}}All pooling functions are\\ replaced by average pooling\end{tabular} & 96.85\% & 53.02\% \\ \cline{2-5}
 & AMTENnet\_8 & \begin{tabular}[c]{@{}c@{}}The 1$\times$1 convolutional kernel\\ in Conv 9 is replaced by 3$\times$3\end{tabular} & 98.36\% & 9.76\% \\ \hline
\end{tabular}
}
\end{table}

\subsection{Design Selection of the AMTENnet Model}
The CNN model has direct impacts on detection accuracy. For the AMTENnet, its AMTEN design is also important, since AMTEN learns manipulation traces for forensics. Actually, AMTEN is very flexible with the following issues to be further investigated by experiments: (1) Are the residual features $F_{mt}$ better for forensics than image data itself? (2) Whether reusing $F_{mt}$ will improve the AMTENnet or not? (3) How many kernels are appropriate for Conv 4 and Conv 5? (4) Is 3$\times$3 convolutional kernel in the first layer better than 5$\times$5 convolutional kernel? (5) What is the effect of the convolution layer on AMTEN?

\begin{figure*}
  \centering
  \includegraphics[width=5.5in]{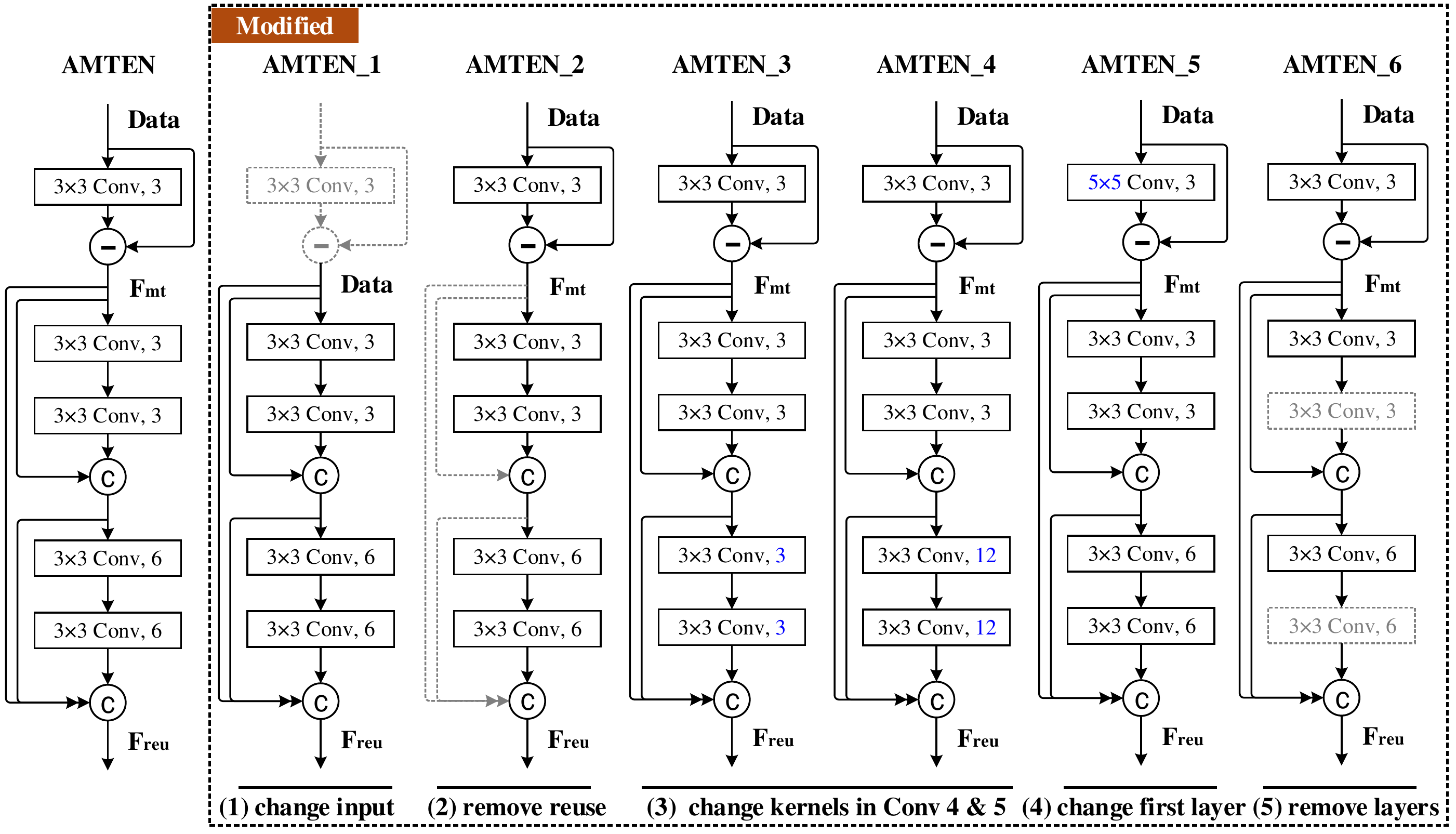}
  \caption{The proposed AMTEN and the six modified versions. The gray dotted line indicates the removed part, and the blue font indicates the modified parameter.}\label{AMTENmodified}
\end{figure*}

To address the above issues, we have made some changes to AMTEN, which are summarized as Fig. \ref{AMTENmodified}. For the AMTENnet model itself, we further discuss two issues: (1) \emph{The pooling layer}. As we know, there are two common pooling strategies, namely max pooling and average pooling. Since AMTENnet adopts max pooling for all the pooling layers, it will be replaced with average pooling for experiments. (2) \emph{The 1$\times$1 convolution layer}. Its main purpose is to achieve cross-channel interaction and information integration. To verify the 1$\times$1 convolutional kernel in Conv 9, it will be replaced with the 3$\times$3 convolutional kernels. To make fair comparisons, We use the same datasets described in Section 4.2 for experiments.

Table \ref{different models} reports the average detection accuracy and $\mathrm{RER}$ for the proposed AMTENnet model with different structures or parameters. From it, we have the following observations. First, AMTEN serves as an effective pre-processing module for the AMTENnet model, since it extracts low-level residual features suitable for forensics. If image data is directly used for feature learning, the detection accuracy will decrease 1.38\%. Second, for the convolutional kernels in the first layer, the size of 3$\times$3 is better than the size of 5$\times$5, which improves the detection accuracy about 0.71\%. A bigger receptive field does not lead to better detection accuracy, and 3$\times$3 convolutional kernels are sufficient for excellent feature extraction. Third, it is a nice choice to use two convolution layers between two concatenation operations. The experimental results prove that two convolution layers are more stable than one convolution layer. Fourth, max pooling is more preferable than average pooling for the AMTENnet, simply because it improves the detection accuracy about 1.67\%. Finally, the 1$\times$1 convolutional kernel in the Conv 9 layer improves 0.16\% detection accuracy than the 3$\times$3 convolutional kernels, which benefits from the cross-channel interaction and information integration.

\begin{table}[]
\centering
\caption{\label{Multiple}Multiple classification identification rate of different forensics models on two datasets.}
\resizebox{90mm}{!}{
\begin{tabular}{|l|c|c|c|c|c|c|c|}
\hline
\multicolumn{1}{|c|}{\multirow{2}{*}{Methods}} & \multicolumn{3}{c|}{HFF dataset}                       & \multirow{2}{*}{Average} & \multicolumn{2}{c|}{FF++ dataset}   & \multirow{2}{*}{Average} \\ \cline{2-4} \cline{6-7}
\multicolumn{1}{|c|}{}                         & Raw              & JP60             & ME5              &                          & HQ               & LQ               &                          \\ \hline
Meso-4                                         & 80.76\%          & 67.76\%          & 62.40\%          & 70.31\%                  & 52.92\%          & 50.63\%          & 51.78\%                  \\ \hline
MesoInception-4                                & 86.40\%          & 58.68\%          & 77.68\%          & 74.25\%                  & 67.38\%          & 45.97\%          & 56.68\%                  \\ \hline
Hand-Crafted-Res                               & 90.54\%          & 73.81\%          & 74.99\%          & 79.78\%                  & 82.94\%          & 64.58\%          & 73.76\%                  \\ \hline
MISLnet                                        & 93.76\%          & 86.32\%          & 79.06\%          & 86.38\%                  & 82.44\%          & 65.61\%          & 74.03\%                  \\ \hline
XceptionNet                                    & 97.17\%          & 78.62\%          & 90.88\%          & 88.89\%                  & 76.30\%          & 71.63\%          & 73.97\%                  \\ \hline
AMTENnet                                       & \textbf{98.52\%} & \textbf{91.02\%} & \textbf{92.42\%} & \textbf{93.99\%}         & \textbf{90.11\%} & \textbf{72.14\%} & \textbf{81.13\%}         \\ \hline
\end{tabular}
}
\end{table}

\begin{table}[]
\centering
\caption{\label{Binary}Binary classification identification rate of different forensics models on two datasets.}
\resizebox{90mm}{!}{
\begin{tabular}{|l|c|c|c|c|c|c|c|}
\hline
\multicolumn{1}{|c|}{\multirow{2}{*}{Methods}} & \multicolumn{3}{c|}{HFF dataset}                       & \multirow{2}{*}{Average} & \multicolumn{2}{c|}{FF++ dataset}   & \multirow{2}{*}{Average} \\ \cline{2-4} \cline{6-7}
\multicolumn{1}{|c|}{}                         & Raw              & JP60             & ME5              &                          & HQ               & LQ               &                          \\ \hline
Meso-4                                         & 76.83\%          & 62.40\%          & 63.52\%          & 67.58\%                  & 61.59\%          & 61.27\%          & 61.43\%                  \\ \hline
MesoInception-4                                & 94.33\%          & 73.63\%          & 84.43\%          & 84.13\%                  & 64.77\%          & 77.21\%          & 70.99\%                  \\ \hline
Hand-Crafted-Res                               & 89.06\%          & 72.97\%          & 76.88\%          & 79.64\%                  & 79.23\%          & 78.79\%          & 79.01\%                  \\ \hline
MISLnet                                        & 93.71\%          & 87.87\%          & \textbf{84.76\%} & 88.78\%                  & 83.84\%          & 82.85\%          & 83.35\%                  \\ \hline
XceptionNet                                    & 92.82\%          & 74.12\%          & 76.62\%          & 81.19\%                  & 75.17\%          & 75.98\%          & 75.58\%                  \\ \hline
AMTENnet                                       & \textbf{97.66\%} & \textbf{88.91\%} & 81.36\%          & \textbf{89.31\%}         & \textbf{85.14\%} & \textbf{84.16\%} & \textbf{84.65\%}         \\ \hline
\end{tabular}
}
\end{table}

\subsection{Comparisons with State-of-the-art works}
To further verify the performance of the AMTENnet model, multiple and binary classification are conducted for detection, respectively. Multiple classification detection is to identify which technique is used to manipulate the face image, which is a fine-grained recognition. Binary classification \footnote{In this task, the fake face images generated by all manipulation methods are classified as fake categories.} is to judge the authenticity of face images, yet it has important advantages for future open-set recognition.

The comparisons are made among AMTENnet and some state-of-the-art works. Note that because Hand-Crafted-Res \cite{binary_highpassfilter} and MISLnet \cite{constrainedCNN} are designed for other forensics tasks, they can not converge for our forensics task. Thus, we replace the initialization method \emph{Gaussian} with \emph{Xavier} \cite{Xavier} for Hand-Crafted-Res, and we adjust the step size to 1000 for MISLnet.

We use two datasets, namely HFF and FF++ dataset, to conduct the experiments. Among them, to hide the traces left by various FIMs, JP and ME are conducted on HFF dataset as post-processing, respectively. The quality factor of JP is set to 60 (JP60), and the kernel size of ME is set to 5$\times$5 (ME5). Table \ref{Multiple} and Table \ref{Binary} report the detection accuracy of multiple and binary classification on two datasets, respectively. We can observe that AMTENnet can achieve the best performance whether it is multiple or binary classification task. Especially in the multiple classification task, AMTENnet outperforms baseline models by a large margin.

\subsection{Detection Robustness}
When face images are spreading over the internet, they usually suffer from some image manipulations with various parameters. It is almost impossible for the detector to learn fake face images under all scenarios. For the AMTENnet, its generalization capability is worthy of further investigation. To further explore the way to improve the detection robustness under complex scenarios, the following two assumptions are made. (1) By applying some post-processing operations to face image data, can the detector learn essential differences among various FIMs to improve its generalization capability? (2) Compared with single parameter, can the post-processing operation with mixed parameters improve the generalization capability of the detector?

\begin{table}[]
\centering
\caption{\label{Parameter}Parameter list of image operations.}
\resizebox{90mm}{!}{
\begin{tabular}{|l|l|l|}
\hline
\multicolumn{2}{|c|}{Image Operations}                           & \multicolumn{1}{c|}{Parameters}                                                                                                                           \\ \hline
\multirow{3}{*}{Spatial Filtering} & Mean Filtering (ME)         & kernel size: 3$\times$3, 5$\times$5, 7$\times$7                                                                                                           \\ \cline{2-3}
                                   & Gaussian Filtering (GB)     & kernel size: 3$\times$3, 5$\times$5, 7$\times$7; Standard deviation: 0                                                                                    \\ \cline{2-3}
                                   & Median Filtering (MED)      & kernel size: 3$\times$3, 5$\times$5, 7$\times$7                                                                                                           \\ \hline
Spatial Enhancement                & Gamma Correction (GC)       & gamma: 0.6, 0.8, 1.0, 1.2, 1.4, 1.6, 1.8, 2.0                                                                                                             \\ \hline
\multirow{2}{*}{Lossy Compression} & JPEG Compression (JP)       & quality factor: 60-90                                                                                                                                     \\ \cline{2-3}
                                   & JPEG Compression 2000 (JP2) & compression ratio: 2.0-8.0                                                                                                                                \\ \hline
Resampling                         & Scaling (SC)                & \begin{tabular}[c]{@{}l@{}}up-sampling (\%): 1, 3, 5, 10, 20, 30, 40, 50, 60, 70, 80, 90\\ down-sampling (\%): 1, 3, 5, 10, 15, 20, 25, 30, 35, 40, 45\end{tabular} \\ \hline
\end{tabular}
}
\end{table}

\begin{table}[]
\centering
\caption{\label{generalization}Confusion matrix for verifying the generalization ability using AMTENnet.}
\resizebox{80mm}{!}{
\begin{tabular}{|c|c|c|c|c|c|c|c|}
\hline
                                                                        & \multicolumn{7}{c|}{Testing Set}                                            \\ \hline
\multirow{6}{*}{\begin{tabular}[c]{@{}c@{}}Training\\ Set\end{tabular}} &        & Raw     & JP60    & JP-mix  & ME5     & ME-mix  & Average          \\ \cline{2-8}
                                                                        & Raw    & \textbf{98.52\%} & 76.41\% & 76.41\% & 75.37\% & 77.11\% & 81.54\% \\ \cline{2-8}
                                                                        & JP60   & 90.19\% & \textbf{91.02\%} & 87.43\% & 54.61\% & 59.40\% & 76.53\%          \\ \cline{2-8}
                                                                        & JP-mix & 93.67\% & 88.78\% & \textbf{90.33\%} & 69.81\% & 73.29\% & 83.18\% \\ \cline{2-8}
                                                                        & ME5    & 74.66\% & 47.57\% & 55.52\% & \textbf{92.42\%} & 87.87\% & 71.61\%          \\ \cline{2-8}
                                                                        & ME-mix & 91.67\% & 65.53\% & 67.45\% & 90.07\% & \textbf{90.74\%} & 81.09\% \\ \hline
\end{tabular}
}
\end{table}

To address the above assumptions, some widely-used image operations are performed on the HFF dataset to simulate face images spreading over the internet. Table \ref{Parameter} summarizes the parameters of these image operations. Note that if the dataset has been suffered from a specific image operation with mixed parameters, it is denoted as `image operation' plus `-mix', such as JP-mix, ME-mix, etc. We select two representative image operations, namely JP and ME, for experiments. Lossy compression can easily confuse the judgment of the detector by reducing image quality. Spatial filtering can hide image details such as manipulation artifacts by blurring the image. They are selected to destroy the traces left in face images by different FIM forgeries. There are five types of face images, which include original face images (Raw), JP60 compressed images (JP60), JP-mix compressed images (JP-mix), ME5 filtered images (ME5), and ME-mix filtered images (ME-mix), respectively. In the experiments, the detector is firstly trained with one type of face images, and then the pre-trained model is tested with the other four types of face images.

Table \ref{generalization} reports the confusion matrixes when AMTENnet are testing five types of face images. We can observe that the detector trained on JP is also effective when detecting Raw and JP, while the detector trained on ME achieves desirable accuracy when detecting Raw and ME. Furthermore, though JP and ME are two distinct image operations to manipulate images, we still observe that the image operations with mixed parameters enable the detector to learn more discriminative features, and thus  improve the generalization ability.

\begin{table}[]
\centering
\caption{\label{mixed data}Identification rate of AMTENnet training on mixed data.}
\resizebox{40mm}{!}{
\begin{tabular}{|c|c|c|c|}
\hline
        & \multicolumn{3}{c|}{Training on mixed data} \\ \hline
        & Small         & Middle       & Large        \\ \hline
Raw     & 95.03\%       & 95.36\%      & 96.32\%      \\ \hline
JP60    & 85.81\%       & 86.46\%      & 88.52\%      \\ \hline
JP-mix  & 86.35\%       & 86.87\%      & 89.31\%      \\ \hline
ME5     & 87.66\%       & 91.92\%      & 91.35\%      \\ \hline
ME-mix  & 88.52\%       & 92.05\%      & 92.82\%      \\ \hline
Average & 88.67\%       & 90.53\%      & 91.66\%      \\ \hline
\end{tabular}
}
\end{table}

We also select face images from Raw, JP-mix and ME-mix datasets with the same proportion to construct Small, Middle and Large mixed datasets, respectively. The AMTENnet model is trained on the mixed training datasets of 124k, 165k, and 372k face images, respectively. Then, the trained detector is used to identify Raw, JP, JP-mix, ME, and ME-mix, respectively. Table \ref{mixed data} reports the experimental results. The detection accuracies also increase with the increase of training data. Their average accuracies are 88.67\%, 90.53\%, and 91.66\%, respectively.

For the generalization capability, there remains a question: whether the detector trained by the above method can detect face images with other unknown operations? To verify this, the trained AMTENnet is tested to detect some other types of face images, such as GB-mix, MED-mix, GC-mix, JP2-mix, and SC-mix. The experimental results are reported in Table \ref{Verify generalization}. The average accuracy is 95.17\%. That is, AMTENnet achieves desirable generalization capability, especially when it is trained on the large dataset. This proves that training the detector with those face images after image operations with mixed parameters is an effective strategy to enhance detection robustness, since the detector can learn more discriminative features from them.

\begin{table}[]
\centering
\caption{\label{Verify generalization}Verify the generalization ability of the AMTENnet.}
\resizebox{55mm}{!}{
\begin{tabular}{|l|c|c|c|c|}
\hline
\multicolumn{2}{|c|}{\multirow{2}{*}{Operation type}} & \multicolumn{3}{c|}{Training on mixed data} \\ \cline{3-5}
\multicolumn{2}{|c|}{}                                & Small         & Middle       & Large        \\ \hline
\multirow{2}{*}{Spatial filtering}      & GB-mix      & 92.07\%       & 94.57\%      & 96.06\%      \\ \cline{2-5}
                                        & MED-mix     & 92.22\%       & 94.75\%      & 95.05\%      \\ \hline
Spatial enhancement                     & GC-mix      & 88.01\%       & 91.89\%      & 93.52\%      \\ \hline
Lossy compression                       & JP2-mix     & 94.64\%       & 94.89\%      & 95.90\%      \\ \hline
Resampling                              & SC-mix      & 94.09\%       & 94.82\%      & 95.34\%      \\ \hline
\multicolumn{2}{|c|}{Average}                         & 92.21\%       & 94.18\%      & 95.17\%      \\ \hline
\end{tabular}
}
\end{table}

\section{Conclusion}
The latest AI-enhanced fake face images have photo-realistic visual qualities, which are quite challenging to be detected.
Due to the relatively fixed structure, there are some limitations for the existing CNN-based works. Thus, we proposed a simple yet effective AMTEN module as pre-processing, which exploits the convolution layer to serve as the predictor to obtain image manipulation traces. The weights are updated adaptively during the back-propagation pass. In subsequent layers, the traces are reused to maximize manipulation traces. We also designed a fake face detector, namely AMTENnet, by integrating AMTEN with CNN.
The manipulation traces obtained by AMTEN are fed into CNN to learn more discriminative features. A series of experiments were conducted, in which several common post-processing operations are selected to simulate the practical forensics under complex scenarios. The experimental results prove that AMTENnet achieved superior detection accuracy and desirable generalization capability. For the HFF dataset, AMTENnet improved average detection accuracy about 7.61\% when compared with MISLnet. Actually, this mainly benefits from AMTEN, since it achieves better residual extraction than Constrained-conv and SRM.
We have also explored the way to improve the detector's robustness. It is worthy of note that AMTEN might also serve as a basic residual predictor for other face forensic tasks.

However, the complex scenarios of practical forensics cases are simulated by several common post-processing operations. Though they do launder the manipulation traces, they are still different from practical cases such as the AI-generated images spreading over social media such as Wechat, Twitter, Facebook and WhatsApp. For future work, we will further improve the robustness of the detector under practical scenarios by gathering real-word samples from social media.

\ifCLASSOPTIONcaptionsoff
  \newpage
\fi


\bibliography{refs}
\bibliographystyle{IEEEtrans}

\end{document}